\documentclass[preprint,12pt]{elsarticle}

\usepackage{amssymb}
\usepackage{amsmath,amsfonts}
\usepackage{algorithmic}
\usepackage{algorithm}
\usepackage{array}
\usepackage[caption=false,font=normalsize,labelfont=sf,textfont=sf]{subfig}
\usepackage{textcomp}
\usepackage{stfloats}
\usepackage{booktabs}
\usepackage{hyperref}
\usepackage{float}
\usepackage{multirow}
\hypersetup{hidelinks}
\usepackage{url}
\usepackage{verbatim}
\usepackage{orcidlink} 
\usepackage{caption}
\usepackage{graphicx}
\DeclareMathAlphabet{\mathbbold}{U}{bbold}{m}{n}
\newcommand{\splitcell}[2][c]{%
  \begin{tabular}[#1]{@{}c@{}}#2\end{tabular}%
}

\makeatletter
\renewcommand\@biblabel[1]{#1. } 
\makeatother %

\journal{Arxiv}

\begin{document}

\begin{frontmatter}



\title{Tackling Small Sample Survival Analysis via Transfer Learning: A Study of Colorectal Cancer Prognosis}


\author[swjtu-leeds]{Yonghao Zhao\orcidlink{009-0007-3180-596X}}
\author[scu]{Changtao Li}
\author[wch-vas]{Chi Shu}
\author[wch]{Qingbin Wu}
\author[scu]{Hong Li}
\author[uestc]{Chuan Xu}
\author[swjtu-ai]{Tianrui Li}
\author[wch]{Ziqiang Wang}

\author[swjtu-ai]{Zhipeng Luo\textsuperscript{*} \orcidlink{0000-0002-4053-5443}}
\author[scu]{Yazhou He\textsuperscript{*}}

\cortext[cor1]{Corresponding authors: 
    Zhipeng Luo (email: zpluo@swjtu.edu.cn), 
    Yazhou He (email: yazhou.he@scu.edu.cn)}
\affiliation[swjtu-leeds]{
    organization={SWJTU-Leeds Joint School, Southwest Jiaotong University},
    city={Chengdu},
    state={Sichuan},
    country={China}}

\affiliation[swjtu-ai]{
    organization={School of Computing and Artificial Intelligence, Southwest Jiaotong University},
    city={Chengdu},
    state={Sichuan},
    country={China}}

\affiliation[scu]{
    organization={Department of Oncology, Division of Epidemiology and Medical Statistics, West China School of Public Health and West China Fourth Hospital, Sichuan University},
    city={Chengdu},
    state={Sichuan},
    country={China}}

\affiliation[wch]{
    organization={Colorectal Cancer Center, Department of General Surgery, West China Hospital, Sichuan University},
    city={Chengdu},
    state={Sichuan},
    country={China}}

\affiliation[wch-vas]{
    organization={Division of Vascular Surgery, Department of General Surgery, West China Hospital, Sichuan University},
    city={Chengdu},
    state={Sichuan},
    country={China}}

\affiliation[uestc]{
    organization={Department of Oncology and Cancer Institute, Sichuan Academy of Medical Sciences, Sichuan Provincial People's Hospital, University of Electronic Science and Technology of China},
    city={Chengdu},
    state={Sichuan},
    country={China}}

\begin{abstract}
Survival prognosis is crucial for medical informatics. Practitioners often confront small-sized clinical data, especially cancer patient cases, which can be insufficient to induce useful patterns for survival predictions.
This study deals with small sample survival analysis by leveraging transfer learning, a useful machine learning technique that can enhance the target analysis with related knowledge pre-learned from other data.
We propose and develop various transfer learning methods designed for common survival models. For parametric models such as DeepSurv, Cox-CC (Cox-based neural networks), and DeepHit (end-to-end deep learning model), we apply standard transfer learning techniques like pretraining and fine-tuning. For non-parametric models such as Random Survival Forest, we propose a new transfer survival forest (TSF) model that transfers tree structures from source tasks and fine-tunes them with target data.
We evaluated the transfer learning methods on colorectal cancer (CRC) prognosis. The source data are 27,379 SEER CRC stage I patients, and the target data are 728 CRC stage I patients from the West China Hospital. When enhanced by transfer learning, Cox-CC's $C^{td}$ value was boosted from 0.7868 to 0.8111, DeepHit's from 0.8085 to 0.8135, DeepSurv's from 0.7722 to 0.8043, and RSF's from 0.7940 to 0.8297 (the highest performance). All models trained with data as small as 50 demonstrated even more significant improvement.
Conclusions: Therefore, the current survival models used for cancer prognosis can be enhanced and improved by properly designed transfer learning techniques. The source code used in this study is available at https://github.com/YonghaoZhao722/TSF.
\end{abstract}








\end{frontmatter}



\section{Introduction}
\label{sec1}
Survival analysis is a fundamental statistical method in medical research, which not only focuses on one or more events of interest, such as death and disease progression, but also the time duration, providing a mixed time-to-event measure for medical research \cite{kleinbaum2005survival}. Common statistical approaches, including the Kaplan-Meier estimator and the Cox proportional hazards model, have enabled researchers to estimate survival probabilities which could be further compared across different treatment groups while adjusting for covariates. This allows for a more nuanced understanding of how prognostic predictors contributed to outcome events over time\cite{cox1972regression}.

The precision and robustness of survival analysis are highly contingent upon the number of patients enrolled and the incidence rate of the event of interest\cite{horiguchi2019accrual}. However, the prolonged follow-up time and relatively low event rates often render it prohibitive to aggregate large sample sizes in prognostic research. For instance, a common study design that investigated prognostic outcomes—clinical trials often involve small sample sizes ($<$200) based on statistics from the FDA\cite{francois2019oncology, ladanie2020clinical}. Furthermore, there has been a growing interest in personalized medicine which relies heavily on analyses in small samples of patients with specific molecular characteristics\cite{hilal2020limitations, downing2014clinical}. These lead to a compelling need in novel analytic approaches that improves efficiency in statistical modelling and prediction performance\cite{schulz2005sample}, in order to help inform clinical decision-making and policy formulation.

Recently, more and more open-access large datasets became available, such as population-based biobanks (e.g. UK Biobank) and registry-based datasets (e.g. Surveillance, Epidemiology, and End Results Program, SEER), providing rich sources of knowledge for studying disease outcomes \cite{palmer2007uk, hankey1999surveillance}. Transfer learning is a powerful technique in machine learning that can leverage these related large datasets to develop a pre-trained model which is then applied to improve learning and prediction performance on a smaller dataset\cite{pan2009survey, cheplygina2019not}.

In this study, we aimed to deal with a common challenge in medical research—to improve prediction performance based on small sample datasets in the setting of survival analysis. We investigated multiple machine learning techniques that were developed for survival analysis, including DeepSurv, Cox-CC, DeepHit, and Random Survival Forests, and developed transfer learning frameworks that incorporated these models to predict survival outcomes. We illustrated these frameworks by providing a real-word example using the SEER database as a pre-trained dataset and applying the trained model to a local cohort of colorectal cancer patients. Specifically, for DeepSurv, Cox-CC, and DeepHit model, we simply applied standard transfer learning techniques including model pretraining (pretrain a network on a source task), retraining (all the network parameters to be re-optimized on the target task), and fine-tuning (part of the parameters to be re-optimized). For random survival forests, we developed novel transferring techniques, which grow a forest on a source task and then “transplant” trees of different depths and of different feature combinations to the target task. 

\section{Background}
In this section, we briefly introduce the basic concepts of survival analysis (SA) and the prevailing SA models including DeepSurv, Cox-CC, DeepHit, and random survival forests. For more in-depth background, please refer to \hyperlink{klein2003survival}{Klein and Moeschberger} (\hyperlink{klein2003survival}{2003}).

In general, the objective of SA is to model the distribution of the time $T^*$ to some event of interest. Denote by $F(t):=P(T^*\le t)$ the cumulative distribution function and by $f(t):= \frac{dF(t)}{dt}$ its density function.
Instead of directly estimating $F(t)$, we often study the survival function $S(t)$ and model the hazard rate $h(t)$, defined as follows:
\begin{equation}
    S(t) := P(T^* > t) = 1 - F(t)
\end{equation}
\begin{equation}
h(t)=\frac{f(t)}{S(t)}=\lim_{\triangle t \to 0}  \frac{1}{\triangle t} P(t\le T^*<t+\triangle t \mid T^*\ge t)
\end{equation}
Then, the cumulative hazard $H(t)$ is frequently used as an intermediate step to calculate $S(t)$:
\begin{equation}
    H(t):=\int_{0}^{t} h(u)du=-\log(S(t))
\end{equation}

\subsection{Cox Regression with Neural Networks}
The Cox Proportional Hazards Model (CPH) is a classic yet popular model in survival analysis. This semi-parametric model assumes that all observations have the same form of the conditional hazard function:
\begin{equation}
    h(t | \mathbf{x})=h_0(t)\exp(g(\mathbf{x})), \hspace{0.5cm} g(\mathbf{x})=\beta^T \mathbf{x}
\end{equation}
where $h_0(t)$ is the baseline hazard function, $\exp[g(\mathbf{x})]$ the relative risk function of given covariates $\mathbf{x}$, and $\beta$ the vector of regression coefficients. 
$\beta$ is usually estimated by minimizing the negative log-partial likelihood: 
\begin{equation}
-l(\beta) = - \sum_{i} \delta_i \log\left( \sum_{j \in R_i}\exp[g(\mathbf{x}_j)-g(\mathbf{x}_i)] \right)
\label{neg_log_par_likelihood}
\end{equation}
where $\delta_i$ is the indicator, being one if the $i$-th individual has experienced the event; and $R_i$ is the set of individuals still at risk at time $t_i$.

The DeepSurv model extends the CPH model by using a neural network to estimate the log-risk function $g(\mathbf{x})$ instead of the linear combination $\beta^T \mathbf{x}$:
\begin{equation}
    \theta_j=G(W\mathbf{x}_j+b)^T\beta
\end{equation}
where $W$ is the weight matrix between the input and hidden layer size $H \times  D$, $H$ the number of neurons in the hidden layer, $D$ the number of input features, $b$ the bias vector, and $G$ a nonlinear activation function. Then, Equation \eqref{neg_log_par_likelihood} can be re-written as:
\begin{equation}
    -l(\beta, W, b)=\sum_{\delta_i=1}{\theta_i -\log\left[ \sum_{j\in R_i} \exp(\theta_j)  \right]}
\end{equation}

To improve the computation efficiency which is important in transfer learning, Kvamme and Borgan (2019)'s Cox-CC (Case-Control) employs a subset $\widetilde{R}_i$ sampled from the full risk set $R_i$ to approximate the entire risk set.

\subsection{DeepHit}
In contrast to Cox-based models, DeepHit by Lee et al. bypasses the Cox hazards and directly estimates the survival distribution $F(t)$ using neural networks.
DeepHit studies on discretized times $0=\tau_0<\tau_1<\tau_2< ... <\tau_m$. Denote by $y_k(\mathbf{x}_i)=\hat{P}(T^*_i=\tau_k\mid \mathbf{x}_i)$ the estimated probability mass function of the times which are outputs of a neural network with covariates $\mathbf{x}$.
Then, the estimated survival function can be written as:
\begin{equation}
    \hat{S}(\tau_k\mid \mathbf{x}_i)=1-\sum^k_{\kappa=1}y_{\kappa}(\mathbf{x}_i)
\end{equation} 
And the discretized negative log-likelihood is defined as:
\begin{equation}
\operatorname{loss}_{L}=-\sum_{i=1}^{N}\left[\delta_{i} \log \left(y_{e_{i}}\left(\mathbf{x}_{i}\right)\right)+\left(1-\delta_{i}\right) \log \left(\hat{S}\left[T_{i} \mid \mathbf{x}_{i}\right]\right)\right]
\end{equation}
where $e_i$ denotes index of time when $T_i=\tau_{e_i}$. 
To further promote the model's ranking ability among individuals, they add a ranking loss defined as:
\begin{equation}
    \operatorname{loss}_R=\sum_{i, j} \delta_{i} \mathbbold{1} \left\{T_{i}<T_{j}\right\} \exp \left(\frac{\hat{S}\left(T_{i} \mid \mathbf{x}_{i}\right)-\hat{S}\left(T_{i} \mid \mathbf{x}_{j}\right)}{\sigma}\right)
\end{equation}
where $\sigma$ is a scale hyper-parameter and $\mathbbold{1}(·)$ the indicator function.
The final loss function combines the above two, i.e., $\operatorname{loss} := \alpha \operatorname{loss}_L+(1-a)\operatorname{loss}_R$, balanced by a hyper-parameter $0< \alpha <1$.

\subsubsection{Random Survival Forests}
Besides parametric models, random survival forests are also popular because of their many advantages, such as good performance, interpretability, and simplicity. Similar to the Classification and Regression Tree (CART), a survival tree is grown with a log-rank splitting rule by reformulating a classification tree \cite{leblanc1993survival,RSF}. Specifically, each terminal node $\mathcal{T}$ of survival trees gives its prediction by fitting individuals $h$ to the Nelson-Aalen estimator:
\begin{equation}
    \hat{H}_h(t)=\sum_{t_{l,h}\le t} \frac{\delta_{l,h}}{Y_{l,h}},\label{nelson}
\end{equation}
where $\delta_{l,h}$ indicates the number of individuals when events occur, and $Y_{l,h}$ is the count of individuals who have not experienced the event up to that time point. 
Random Survival Forests (RSFs) is an ensemble of $N$ survival trees with bootstrap resampling. Aggregating their predictions of \eqref{nelson}, the cumulative hazard function of the bootstrap ensemble of a forest for an individual $i$ is computed as:
\begin{equation}
    H(t\mid \mathbf{x}_i)=\frac{1}{N} \sum^N_{n=1} H_n(t\mid \mathbf{x}_i).
\end{equation}
where $n$ means the $n$-th tree. 

\section{Methods}

In this section, we detail our proposed transfer learning methods for different SA models. There are two categories - one works for neural network based (parametric) models, and the other for random survival forests (non-parametric). Afterward, we will present the experimental procedure to validate our methods.

\subsection{Transfer for Neural Networks}

Current transfer learning techniques designed for neural networks are relatively well-developed. 
The foundation of transfer learning is that the source task and the target task should be somehow related such that the target can benefit from the knowledge learned from the source.
The typical process is that a model pretrained on a source task is retrained or fine-tuned on the target.
Retraining (RT) allows all the pretrained model parameters to be re-optimized in the target task such that the model can be well-adapted; by comparison, fine-tuning (FT) only allows part of the parameters to be changed such that the pre-learned knowledge can be selectively preserved. For deep neural nets, usually the deep layers are frozen such that low-level or general feature representations are transferred for reuse\cite{freezelayer}; then the last few layers are fine-tuned such that the model can adapt to task-specific signals\cite{ha2024domain}.

In our study, DeepSurv, Cox-CC, and DeepHit are all neural network based models and can readily work with the retraining and fine-tuning techniques. A particular aspect of our study is that we work on small samples. So, choosing what granularity of transfer (i.e. which parameters to preserve and which to re-optimize) depends on the data availability of the target task. On the one hand, fine-tuning can be viewed as the generalization of retraining, as retraining is one extreme case of fine-tuning; on the other hand, from the perspective of parameter optimization, if provided with sufficient data, retraining can achieve more optimal results than fine-tuning, since the latter only optimizes part of the parameters. However, when encountering severe data limits, retraining may cause inferior transfer outcomes than fine-turning \cite{finetune-layer}. Therefore, we study the relationship between transfer granularity and data availability. In particular, we limit our fine-tuning to only adjusting the output layer, which is another extreme case. We empirically show that when the target data are relatively abundant (when $\ge 500$ cases), retraining leads to better transfer performance; when the data are limited (when $\le 200$ cases), fine-tuning outperforms retraining. See \hyperref[tab:nn_res]{Table~\ref*{tab:nn_res}} for the detailed results.

\begin{figure*}[htbp]
    \centering
    \includegraphics[width=\linewidth]{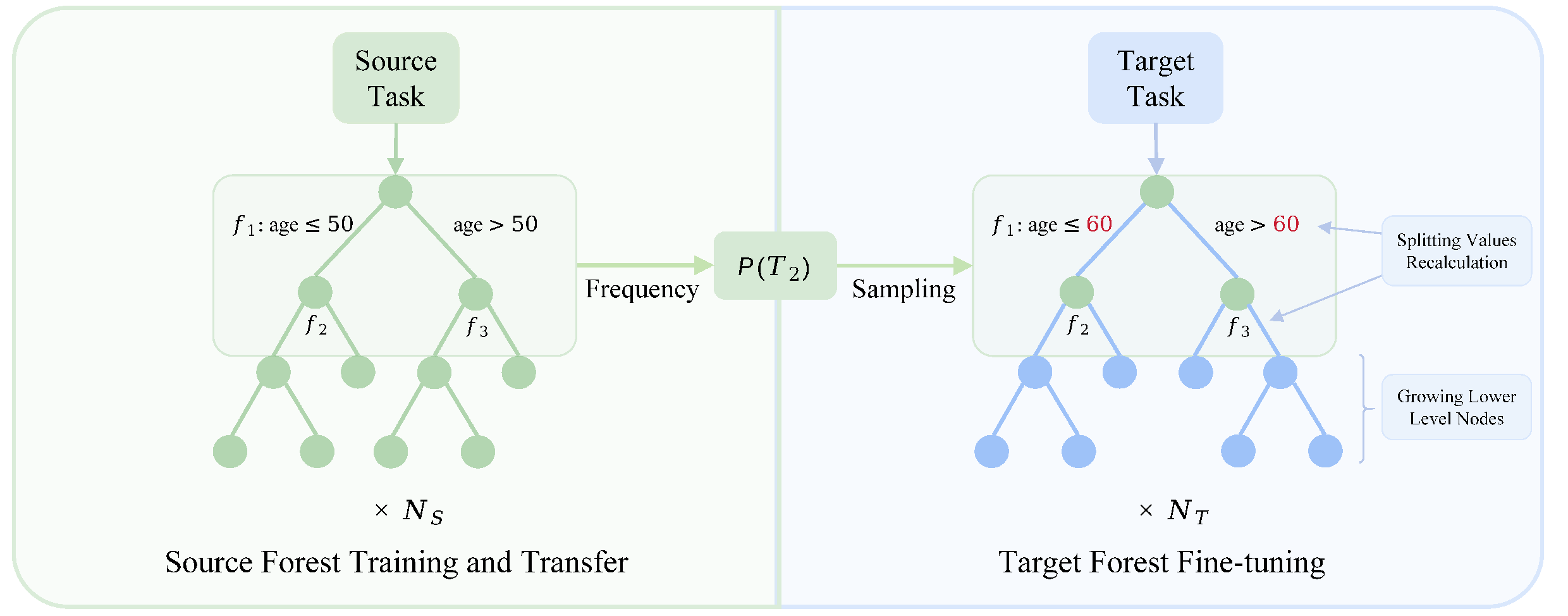} 
    \caption{
    An illustration of using the TSF-$T_2$ transfer method to build a random survival forest. Left is to train a source forest of $N_S$ trees and compute the empirical probability of two-level tree structures $P(T_2)$. Right is to randomly sample $N_T$ prototype trees based on $P(T_2)$ and perform fine-tuning. The top two levels use the same splitting features but recalculate the splitting values based on the target data; the lower-level trees are growing in a standard way, independently of the transferred information.
    }
    \label{fig:tsfflow}
\end{figure*}

\subsection{Transfer for Random Survival Forests}
\label{subsec:tsf}

Now we present the transfer techniques for random survival forests (RSF), which are the main contributions of our study. 
The general workflow is to first build an RSF (called the \textit{source forest}) based on the source task and then randomly transfer the frequent tree structures to the target task, where a \textquotedblleft fine-tuning\textquotedblright\ procedure is performed. Tree structures differ in their features used to split nodes on different levels, and the tree-based fine-tuning is to adjust the splitting values and/or to grow lower-level nodes. The idea behind our method is intuitive - similar tasks are likely to generate similar survival tree structures, which can thus be transferred; on the other hand, to adapt to the target task, the splitting values of the transferred structures can be re-optimized, and new lower-level nodes can be further grown.

\subsubsection{Source Forest Training and Transfer}

We start with calling a standard procedure \cite{leblanc1993survival,RSF} to train a random survival forest based on the source task's data. This gives us $N_S$ survival trees, where each tree differs in its tree structures and splitting values. Though randomly built, these trees can reflect the most discriminative features that determine the survival outcomes. In other words, discriminative features tend to appear more frequently and on higher levels of the trees. Therefore, key to the tree-based transfer comes down to the frequency and the level of occurrence of features. Meanwhile, the inherent cross-feature relationship also matters, which is reflected in the co-occurrence of features on the same or adjacent levels.
In such spirit, we design a transfer principle named tree-structure frequency (TSF-$T_k$) that counts the frequency of different tree structures with top $k$ levels in the $N_S$ trees. 
The frequencies can be normalized to a probability distribution $P(T_k)$, which can be used to sample $N_T$ prototype trees with $k$ levels that are further fine-tuned with the target task's data. 
See \hyperref[fig:tsfflow]{Figure \ref{fig:tsfflow}} for an example.


Besides TSF, we also develop a comparative transfer method named DP that only counts the \textbf{d}epth-wise \textbf{p}robability of feature occurrence. Formally, DP outputs a set of probabilities $P_{DP} = \{P_1,P_2,\cdots,P_K\}$, where $P_k(f)$ denotes the empirical probability distribution of feature $f$ that occurred on the $k$-th level of the source trees.
Different from TSF, DP only considers feature occurrences \textit{independently} for each level, ignoring the co-occurrence of other features on adjacent levels. Therefore, when used to build new trees for the target task, DP can explore new tree structures while down-weighting existing structures occurring in the source forest.

\subsubsection{Target Forest Fine-tuning}

Now we build a target forest of $N_T$ trees based on the transferred information. TSF-$T_k$ outputs an empirical probability distribution of tree structures $P(T_k)$, and DP outputs a depth-wise feature distribution $P_{DP}$. Below is the forest building process based on TSF-$T_k$:
\begin{enumerate}
    \item Sample a tree structure $T$ of $k$ levels according to $P(T_k)$.
    \item Build a tree $T'$, of which the top $k$ levels have the same splitting features as $T$ but different splitting values which are determined by the target training data; for lower levels $>k$, follow a standard tree-building process to determine the splitting features and values. See \hyperref[fig:tsfflow]{Figure \ref{fig:tsfflow}} for an illustration.
    \item Repeat the above procedure independently until $N_T$ trees are built.
\end{enumerate}
The DP-based method works similarly, except that the splitting feature of each node on level $k$ is randomly chosen by $P_k(f)$. The splitting values are, again, determined only by the target training data.

\section{Experiments and Results}

In this section, we evaluate our proposed transfer learning methods for different survival models. The target learning task is to predict survival outcomes of stage I colorectal cancer patients (CRC) collected from the West China Hospital (WCH)\cite{36600277}. The source task's data are collected from the Surveillance, Epidemiology, and End Results (SEER) database\cite{nihseer}, of which CRC stage I patients are used. Below we present the data statistics, experiment setups, and results.

\subsection{Cohort Statistics}
SEER and WCH respectively have 27,379 and 728 CRC stage I patient cases. Eight features are used as predictors: gender, age, T stage (T1 or T2), tumor size, tumor grade (G3-G4 or G1-G2), Carcinoembryonic Antigen (CEA) levels (positive when $>$2.5 ng/mL, negative, or unknown), perineural invasion (positive or negative), and suboptimal lymph node examination (lymph nodes sampling $>$12 or not).
These covariates have been shown to be strong indicators in the prognosis of stage I CRC patients \cite{36600277}
For feature-wise statistics, SEER has 51.1$\%$ males while WCH has 55.7$\%$ males; SEER patients have a median age of 69 years old whereas the median age in WCH is 61 years old. Clinical features show more variations. For example, SEER has a median tumor size of 2.4 and WCH's is 3; for CEA levels, SEER's positive rate is 89.8$\%$ while WCH's is 52.2$\%$. Besides, the mortality rates also differ significantly, with SEER's being 25.4$\%$ and WCH's being 5.3$\%$, which is highly censored.

\subsection{Experimental Design}

Firstly, the SEER dataset is used to pretrain models. We use cross-validation to optimize the pretrained models.
For the target task, the full WCH dataset has 728 samples, which seems sufficient for survival analysis. However, in practice, the available cancer cases can be extremely limited. To test with small samples, we experiment with different sample sizes $n$, which gradually shrink from 500, 200, 100, and 50 to $<$50. This setup allows us to observe how transfer learning methods' performance varies with sample size. To fairly compare the results of different sample sizes, we use a universal 10-fold stratified cross-validation setup. Specifically, first, the full 728 samples are equally divided into ten folds. Then for each run of the cross-validation, one fold of data is held for test, and the remaining nine folds are used to sample $n$ training data, used for retraining or fine-tuning. This ensures that all results are generated based on the same test data.
We employed the time-dependent concordance index $C^{td}$\cite{ctd} as the evaluation metric. $C^{td}$ measures the agreement between the ordering of predicted risk and the actual survival times of pairs, and is widely used for assessing survival models \cite{2024concordance}. 
The \textit{pycox} library nicely provides the implementation of DeepSurv, Cox-CC, and DeepHit models. We implemented our proposed Transfer Survival Forest (TSF) method, available at Github\footnote{\href{https://github.com/YonghaoZhao722/TSF}{https://github.com/YonghaoZhao722/TSF}}.

\subsection{Transfer Learning Results}

\begin{table}[htbp]
\centering
\resizebox{\textwidth}{!}{

\begin{tabular}{cccccc}
\toprule
Data Size & Model & Target & Source & FT & RT \\
\midrule

\multirow{3}{*}{\splitcell{WCH 655\\\footnotesize{(Full)}}} & Cox-CC& 0.7868±0.0990 & 0.8008±0.1829 & 0.8039±0.1343 & \textbf{0.8111±0.1383} \\
                         & DeepHit & 0.8085±0.1507 & 0.8116±0.1492 & 0.8124±0.1589 & \textbf{0.8135±0.1578} \\
                         & DeepSurv & 0.7722±0.1373 & 0.7878±0.1807 & \textbf{0.8043±0.1515} & 0.8018±0.1531 \\
\midrule
\multirow{3}{*}{WCH 500} & Cox-CC& 0.7652±0.1535 & 0.8008±0.1829 & 0.8061±0.1566 & \textbf{0.8128±0.1326} \\
                         & DeepHit & 0.7682±0.1255 & 0.8116±0.1492 & 0.8008±0.1548 & \textbf{0.8047±0.1357} \\
                         & DeepSurv & 0.7359±0.1095 & 0.7878±0.1807 & \textbf{0.7976±0.1732} & 0.7940±0.1671 \\
\midrule
\multirow{3}{*}{WCH 200} & Cox-CC& 0.7408±0.1355 & 0.8008±0.1829 & 0.8055±0.1165 & \textbf{0.8081±0.1501} \\
                         & DeepHit & 0.7352±0.1627 & 0.8116±0.1492 & \textbf{0.8114±0.1577} & 0.8046±0.1605 \\
                         & DeepSurv & 0.7228±0.1449 & 0.7878±0.1807 & \textbf{0.7974±0.1714} & 0.7949±0.1509 \\
\midrule
\multirow{3}{*}{WCH 100} & Cox-CC& 0.6961±0.1963 & 0.8008±0.1829 & 0.8067±0.1706 & \textbf{0.8084±0.1672} \\
                         & DeepHit & 0.7330±0.1691 & 0.8116±0.1492 & \textbf{0.8107±0.1601} & 0.7954±0.1773 \\
                         & DeepSurv & 0.6541±0.2006 & 0.7878±0.1807 & \textbf{0.7905±0.1641} & 0.7854±0.1736 \\
\midrule
\multirow{3}{*}{WCH 50}  & Cox-CC& 0.6252±0.2155 & 0.8008±0.1829 & \textbf{0.8027±0.1719} & 0.8004±0.1713 \\
                         & DeepHit & 0.6334±0.1542 & 0.8116±0.1492 & \textbf{0.8064±0.1591} & 0.7874±0.1486 \\
                         & DeepSurv & 0.6150±0.1679 & 0.7878±0.1807 & \textbf{0.7959±0.1725} & 0.7864±0.1787 \\
\bottomrule
\end{tabular}

}
\caption{Transfer learning results ($C^{td}$ scores) for parametric survival models with different training data sizes. \textit{Target}: only using the target data, no transfer; \textit{Source}: directly using the pretrained SEER model, no fine-tuning or retraining; \textit{FT}: fine-tuning; \textit{RT}: retraining. Each row's highest $C^{td}$ score (except \textit{Source}) is marked in bold.}
\label{tab:nn_res}
\end{table}

The first set of results focuses on the transfer learning performance for parametric models, i.e. Cox-based models and neural networks. \hyperref[tab:nn_res]{Table~\ref*{tab:nn_res}} lists the retraining (RT) and fine-tuning (FT) results when tested with different models and different data sizes. We also provide two baselines for comparison. One is given by the \textit{Target} column, which means using only the target data for training, with zero transfer. The other is given by the \textit{Source} column, which means directly applying the pretrained SEER model on the test data, with zero fine-tuning. Hence, this column remains unchanged for different sample sizes.

\begin{table}[!htbp]
    \centering
    \resizebox{\textwidth}{!}{

\begin{tabular}{ccccccccc} 
    \toprule
    \multirow{2}[0]{*}{Data Size} & Target& \multirow{2}[0]{*}{Source} & \multicolumn{5}{c}{TSF-$T_k$} & \multirow{2}[0]{*}{DP-Based} \\ 
    \cmidrule(lr){4-8}
          &       (RSF)&       & $T_1$    & $T_2$    & $T_3$    & $T_4$    & $T_\infty$    &  \\
    \midrule

\splitcell{WCH 655\\\footnotesize{(Full)}} & \splitcell{0.7940\\\footnotesize{$\pm$0.1922}} & \splitcell{0.7994\\\footnotesize{$\pm$0.1629}} & \splitcell{0.8232\\\footnotesize{$\pm$0.1193}} & \splitcell{0.8151\\\footnotesize{$\pm$0.1265}} & \textbf{\splitcell{0.8297\\\footnotesize{$\pm$0.1098}}} & \splitcell{0.8141\\\footnotesize{$\pm$0.1245}} & \splitcell{0.8128\\\footnotesize{$\pm$0.1157}} & \splitcell{0.8089\\\footnotesize{$\pm$0.1312}} \\
\midrule
WCH 500 & \splitcell{0.7791\\\footnotesize{$\pm$0.1224}} & \splitcell{0.7994\\\footnotesize{$\pm$0.1629}} & \textbf{\splitcell{0.8111\\\footnotesize{$\pm$0.1529}}} & \splitcell{0.8040\\\footnotesize{$\pm$0.1377}} & \splitcell{0.7943\\\footnotesize{$\pm$0.1153}} & \splitcell{0.7913\\\footnotesize{$\pm$0.1168}} & \splitcell{0.7969\\\footnotesize{$\pm$0.1490}} & \splitcell{0.7870\\\footnotesize{$\pm$0.1660}} \\ 
\midrule
WCH 200 & \splitcell{0.7612\\\footnotesize{$\pm$0.1955}} & \splitcell{0.7994\\\footnotesize{$\pm$0.1629}} & \textbf{\splitcell{0.8044\\\footnotesize{$\pm$0.1630}}} & \splitcell{0.7978\\\footnotesize{$\pm$0.1503}} & \splitcell{0.7950\\\footnotesize{$\pm$0.1529}} & \splitcell{0.7955\\\footnotesize{$\pm$0.1534}} & \splitcell{0.7951\\\footnotesize{$\pm$0.1545}} & \splitcell{0.7719\\\footnotesize{$\pm$0.1956}} \\ 
\midrule
WCH 100 & \splitcell{0.7578\\\footnotesize{$\pm$0.1742}} & \splitcell{0.7994\\\footnotesize{$\pm$0.1629}} & \textbf{\splitcell{0.7960\\\footnotesize{$\pm$0.1688}}} & \splitcell{0.7748\\\footnotesize{$\pm$0.1676}} & \splitcell{0.7809\\\footnotesize{$\pm$0.1572}} & \splitcell{0.7809\\\footnotesize{$\pm$0.1572}} & \splitcell{0.7809\\\footnotesize{$\pm$0.1572}} & \splitcell{0.7434*\\\footnotesize{$\pm$0.1945}}\\ 
\midrule
WCH 90 & \splitcell{0.7417\\\footnotesize{$\pm$0.1586}} & \splitcell{0.7994\\\footnotesize{$\pm$0.1629}} & \textbf{\splitcell{0.7857\\\footnotesize{$\pm$0.1444}}} & \splitcell{0.7683\\\footnotesize{$\pm$0.1544}} & \splitcell{0.7809\\\footnotesize{$\pm$0.1338}} & \splitcell{0.7819\\\footnotesize{$\pm$0.1338}} & \splitcell{0.7819\\\footnotesize{$\pm$0.1471}} & \splitcell{0.7300*\\\footnotesize{$\pm$0.1814}}\\ 
\midrule
WCH 80 & \splitcell{0.7393\\\footnotesize{$\pm$0.1421}} & \splitcell{0.7994\\\footnotesize{$\pm$0.1629}} & \textbf{\splitcell{0.8050\\\footnotesize{$\pm$0.1266}}} & \splitcell{0.7757\\\footnotesize{$\pm$0.1333}} & \splitcell{0.7851\\\footnotesize{$\pm$0.1562}} & \splitcell{0.7851\\\footnotesize{$\pm$0.1407}} & \splitcell{0.7851\\\footnotesize{$\pm$0.1407}} & \splitcell{0.7429\\\footnotesize{$\pm$0.1848}} \\ 
\midrule
WCH 70 & \splitcell{0.7310\\\footnotesize{$\pm$0.1371}} & \splitcell{0.7994\\\footnotesize{$\pm$0.1629}} & \splitcell{0.7456\\\footnotesize{$\pm$0.1350}} & \textbf{\splitcell{0.7721\\\footnotesize{$\pm$0.1656}}} & \splitcell{0.7673\\\footnotesize{$\pm$0.1635}} & \splitcell{0.7673\\\footnotesize{$\pm$0.1635}} & \splitcell{0.7673\\\footnotesize{$\pm$0.1635}} & \splitcell{0.7322\\\footnotesize{$\pm$0.1481}} \\ 
\midrule
WCH 60 & \splitcell{0.7380\\\footnotesize{$\pm$0.1168}} & \splitcell{0.7994\\\footnotesize{$\pm$0.1629}} & \splitcell{0.7487\\\footnotesize{$\pm$0.1279}} & \textbf{\splitcell{0.7777\\\footnotesize{$\pm$0.1422}}} & \splitcell{0.7721\\\footnotesize{$\pm$0.1458}} & \splitcell{0.7721\\\footnotesize{$\pm$0.1458}} & \splitcell{0.7721\\\footnotesize{$\pm$0.1458}} & \splitcell{0.7360*\\\footnotesize{$\pm$0.1305}}\\ 
\midrule

WCH 50 & \splitcell{0.7267\\\footnotesize{$\pm$0.1176}}& \splitcell{0.7994\\\footnotesize{$\pm$0.1629}} & \textbf{\splitcell{0.7498\\\footnotesize{$\pm$0.1816}}}& \splitcell{0.7381\\\footnotesize{$\pm$0.1996}}& \splitcell{0.7082*\\\footnotesize{$\pm$0.2207}}& \splitcell{0.7067*\\\footnotesize{$\pm$0.2173}}& \splitcell{0.7067*\\\footnotesize{$\pm$0.2172}}& \splitcell{0.7351\\\footnotesize{$\pm$0.1473}}\\ 
\midrule

WCH 40 & \splitcell{0.7329\\\footnotesize{$\pm$0.1326}} & \splitcell{0.7994\\\footnotesize{$\pm$0.1629}} & \splitcell{0.7172*\\\footnotesize{$\pm$0.1616}}& \splitcell{0.7382\\\footnotesize{$\pm$0.1831}} & \splitcell{0.7121*\\\footnotesize{$\pm$0.2242}}& \splitcell{0.7121*\\\footnotesize{$\pm$0.2242}}& \splitcell{0.7121*\\\footnotesize{$\pm$0.2242}}& \textbf{\splitcell{0.7562\\\footnotesize{$\pm$0.1702}}} \\ 
\midrule

WCH 30 & \textbf{\splitcell{0.7100\\\footnotesize{$\pm$0.1146}}} & \splitcell{0.7994\\\footnotesize{$\pm$0.1629}} & \splitcell{0.6669*\\\footnotesize{$\pm$0.2022}}& \splitcell{0.6825*\\\footnotesize{$\pm$0.1915}}& \splitcell{0.6825*\\\footnotesize{$\pm$0.1915}}& \splitcell{0.6825*\\\footnotesize{$\pm$0.1915}}& \splitcell{0.6825*\\\footnotesize{$\pm$0.1915}}& \splitcell{0.6694*\\\footnotesize{$\pm$0.1395}}\\ 
\midrule

WCH 20 & \textbf{\splitcell{0.6607\\\footnotesize{$\pm$0.1753}}} & \splitcell{0.7994\\\footnotesize{$\pm$0.1629}} & \splitcell{0.5659*\\\footnotesize{$\pm$0.1213}}& \splitcell{0.5697*\\\footnotesize{$\pm$0.1897}}& \splitcell{0.5697*\\\footnotesize{$\pm$0.1897}}& \splitcell{0.5697*\\\footnotesize{$\pm$0.1897}}& \splitcell{0.5697*\\\footnotesize{$\pm$0.1897}}& \splitcell{0.6485*\\\footnotesize{$\pm$0.1900}}\\ 
\bottomrule

\end{tabular}%
    }
    \caption{Transfer learning results ($C^{td}$ scores) for transfer survival forest (TSF) with different training data sizes. \textit{Target}: only using the target data to train a random survival forest (RSF), no transfer; \textit{Source}: directly using the pretrained RSF based on SEER, no fine-tuning; TSF-$T_k$ and DP-based are different configurations of TSF. Each row's highest $C^{td}$ score (except \textit{Source}) is marked in bold. Values lower than \textit{Target} are asterisked.}
    \label{tab:rsf_res}
\end{table}

The second set of results is for transfer survival forest (TSF), presented in \hyperref[tab:rsf_res]{Table }\ref{tab:rsf_res}. Similarly, the \textit{Target} column means no transfer and just using a normal random survival forest (RSF); and the \textit{Source} column means no fine-tuning and directly using a pretrained RSF based on the SEER data. The remaining columns are the results of different TSF variations (see \hyperref[subsec:tsf]{Methods}). Note that TSF-$T_k$ refers to transferring and fine-tuning only the top $k$ levels of trees, and $T_\infty$ means no such level limitation.

\section{Discussions}

In the preceding section, we empirically evaluated various transfer learning methods designed for parametric survival models (DeepSurv, Cox-CC, DeepHit) and random survival forests. The base learning task was CRC prognosis, using SEER as the source and WCH as the target, and different sample sizes were compared. Here, we are to analyze these empirical results and reveal the insights behind them. Also, limitations and future work will be discussed as well.

\subsection{Clinical Implications}
Our findings demonstrate that transfer learning significantly enhances the predictive accuracy of survival analysis based on limited sample sizes. This finding is of important clinical relevance considering the prohibitive cost and duration of developing large-scale patient cohorts that request prolonged follow-up and that investigate rare disease outcomes.  
Although our case-study utilized common cinico-pathological features of cancer patients, the framework we propose can be applied to other features such as imaging and molecular omics data provided by large international biobanks. In additon to cancer research, these techniques are generally applicable to common disease outcomes measured in a time-to-event manner, such as investigations in outcomes of critical care using electronic health records (EHR) data. 

Recent studies have demonstrated the utility of deep learning approaches for predicting patient outcomes. For example, researchers developed a workflow that combines convolutional neural networks (CNNs) with transfer learning strategies to overcome overfitting issues in scenarios with limited training datasets composed of high-dimensional samples, such as those encountered in cancer predictive tasks involving gene-expression data\cite{lopez2020transfer}. This approach has been extended by several studies that employed transfer learning techniques for survival analysis models. Specifically, Transfer-Cox leveraged features learned from source tasks to enhance a basic Cox model\cite{li2016transfer}, while Zhu et al\cite{deepsurv_transfer}. and Gao et al[16]. adapted DeepSurv through pre-training on large datasets followed by retraining with target datasets. TLSurv utilized multi-modal genomic data, first extracting features using multiple neural networks and then applying them to a Cox model\cite{tlsurv}. Bellot and van der Schaar proposed a boosting method for growing survival trees that can be transferred across domains\cite{bellot2019boosting}. Notably, these studies primarily focused on basic transfer learning techniques applied to parametric models in specific areas, thus limiting their broader applicability. A key gap remains in extending these methods to tree-based models, which are increasingly popular due to their interpretability and effectiveness in modeling complex survival data. Our study addressed this gap by deploying a transfer learning method for random survival forests.

\subsection{Empirical Findings}

Firstly, for neural network models (\hyperref[tab:nn_res]{Table~\ref*{tab:nn_res}}), we can find that both fine-tuning and retraining can boost all the models' performance on different sample sizes (as small as fifty). The first reason accounting for this is that the raw transferred model (the Source column) can provide a strong baseline superior to the non-transfer model (the Target column).

This demonstrates a key principle of transfer learning
 - indeed, the source and target tasks should be somewhat related such that the learned knowledge, represented by models or in other forms, can be transferred and benefit the target. Moreover, when the source model is fine-tuned or retrained with the target data, the model performance can be further improved. This holds for all the cases of Cox-CC, DeepSurv, and partially for DeepHit (yet the inferior cases are not behind too much). Such evidence demonstrates the efficacy of transfer learning.
Besides, when comparing fine-tuning and retraining, we mentioned earlier that from the perspective of optimization, fine-tuning is a special case of retraining, so theoretically the latter should always outperform. However, when data are limited, fine-tuning may be favored since insufficient data can hardly help in finding more optimal results. This is evidenced in \hyperref[tab:nn_res]{Table~\ref*{tab:nn_res}} - when data are relatively rich ($\ge$500), retraining surpasses fine-tuning; when data volumes drop till fifty, the situation gradually reverses. Therefore, the practical guide is that choosing fine-tuning or retraining or what extent of fine-tuning depends on the data availability. Such a finding reflects and extends the work of Zhu et al. \cite{deepsurv_transfer}.

\begin{figure*}[htbp]
    \centering
    \includegraphics[width=\linewidth]{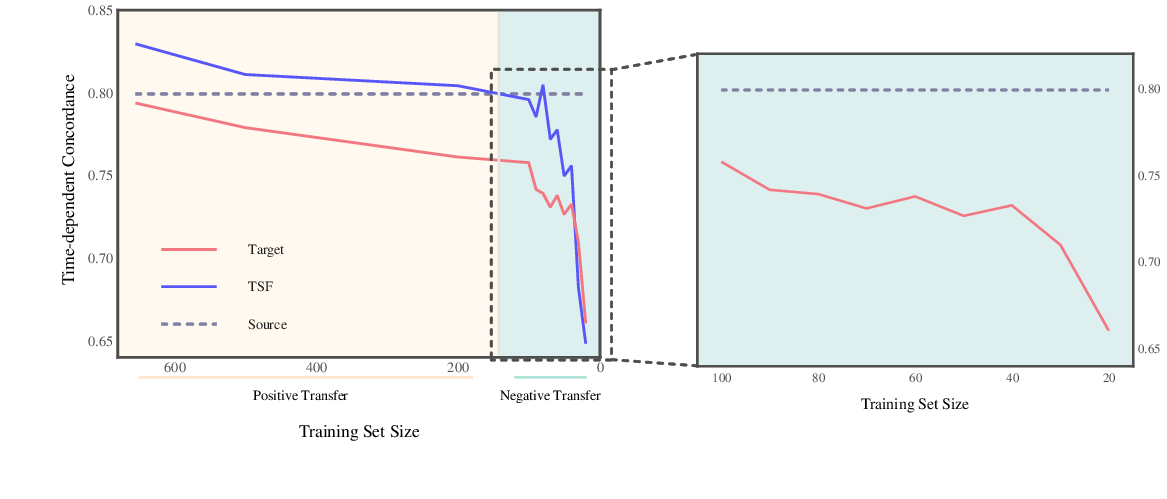} 
    \vspace{-30pt}
    \caption{A trend plot showing how TSF's performance varies with different training sizes, compared with the target model (zero transfer) and the source model (zero fine-tuning). The TSF curve represents the highest $C^{td}$ value among all the variations. The boundary of positive and negative transfer lies where TSF starts to perform worse than the source model.}
    \label{fig:tsfres}
\end{figure*}

Regarding transfer survival trees (TSF), in \hyperref[tab:rsf_res]{{Table }}\ref{tab:rsf_res}, we can find that TSFs can outperform non-transfer models when training data sizes are $\ge$40. Remarkably, TSFs outputted the highest $C^{td}$ scores ($>$0.82) among all models in all experiments.
The raw source model still provides a strong baseline better than the target model. When the target training data are relatively sufficient ($\ge$180, \hyperref[fig:tsfres]{{Figure~\ref*{fig:tsfres}}}), fine-tuned forests can achieve better predictive performance. This is termed \textit{postive transfer} as in the transfer learning community \cite{survey_transfer}. We also see negative transfer occur, which becomes notable when the data size drops to $<$80. The main reason is due to the splitting value recalculation in survival trees. Recall that TSF only transfers feature combinations of the source trees, and the splitting values are redefined by the target data. This largely differs from parametric models in that TSF still relies more on the target data. Consequently, limited volumes of data suffer from high variance and are less likely to provide generalizable splitting values.
This also explains why TSF performs even worse than the non-transfer models when the training data are severely limited ($<$40). Therefore, in such situations, we make conservative recommendations, that only transfer the top one or two levels of features as TSF-$T_1$ and $T_2$; that avoid building overly deep trees to mitigate over-fitting (in our experiments, within three levels is suggested).

\subsection{Limitations and Future Work}

While our study has demonstrated the feasibility and benefits of applying transfer learning to cancer prognosis, there are certain limitations and several promising future directions to discuss. 
First, the SEER database was able to provide strong source models to be transferred, which, however, downplayed the effect of the following modifications of the source models. It would be interesting to see how much a mild source model could contribute to the target and how much margin the fine-tuning techniques with limited target data could improve the source model.
Second, the source and the target data we experimented with have completely aligned feature sets. However, many related clinical tasks' feature sets differ, and how to handle non-overlapped features is a critical challenge in transfer learning.
Last but not least, nowadays with richer data collections of genomics and proteomics data, it will be of tremendous interest to leverage multi-modal data for cancer prognosis. For example, combine pre-trained cell foundation models (such as GeneFormer \cite{Theodoris2023transfer}) and bulk RNA-seq data with survival observations and then design comprehensive and more precise survival models.


\section{Conclusions}

Our study aims to tackle small sample survival analysis with the help of transfer learning. We summarized the current transfer learning methods applied to trending neural network based survival models (DeepSurv, Cox-CC, DeepHit) and proposed new transfer techniques for random survival forests. We also evaluated these methods on a set of colorectal cancer prognosis tasks, where the source data are from SEER and the target data are collected from the West China Hospital (WCH). 
Based on the experiment results, the two sets of transfer methods are shown to improve all the non-transfer models on various small sample sizes. This demonstrates the feasibility and superiority of using transfer learning to cancer prognosis.
Comparing neural network survival models and random survival forests (RSFs), the former appears to be more robust against small sample sizes as the transfer process relies less on the target data; in contrast, RSFs have better interpretability but rely more on data for fine-tuning, and thus it is recommended to build shallower trees (to avoid over-fitting) and make minor modifications (to avoid under-fitting).


\section{Acknowledgments}
The work presented was supported by the National Natural Science Foundation of China (No. 62302405, 62176221, 82103918), China Postdoctoral Science Foundation (Grant No. 2023M732914), and the Fundamental Research Funds for the Central Universities (No. 2682023ZT007).
The content of this paper is solely the responsibility of the authors and does not necessarily represent the official views of the abovementioned funding agencies.

\section{Ethics approval and consent to participate}
This study was approved by the Ethics Committee of West China Hospital(approval lD: 201960V1.0 for the WCH cohort).Written informed consent was signed by each participant from the study cohorts.

\section{Source Availability}
All data are available upon reasonable request; please refer to the details elsewhere \cite{36600277}. The source code for the experimental implementation used in this study is available at the Github repository (\href{https://github.com/YonghaoZhao722/TSF}{\underline{https://github.com/YonghaoZhao722/TSF}}).

\bibliographystyle{elsarticle-num} 
\bibliography{citation}


\end{document}